\documentclass[sigconf]{acmart}
\usepackage[linesnumbered,ruled,vlined]{algorithm2e}
\usepackage{booktabs,makecell,multirow}
\usepackage{multirow}
\AtBeginDocument{%
  \providecommand\BibTeX{{%
    \normalfont B\kern-0.5em{\scshape i\kern-0.25em b}\kern-0.8em\TeX}}}

\copyrightyear{2023}
\acmYear{2023}
\setcopyright{rightsretained}
\acmConference[GECCO '23 Companion]{Genetic and Evolutionary Computation Conference Companion}{July 15--19, 2023}{Lisbon, Portugal}
\acmBooktitle{Genetic and Evolutionary Computation Conference Companion (GECCO '23 Companion), July 15--19, 2023, Lisbon, Portugal}\acmDOI{10.1145/3583133.3590735}
\acmISBN{979-8-4007-0120-7/23/07}

\begin{document}

\title{A Transformer-based Neural Architecture Search Method}

\author{Shang Wang}
\affiliation{%
  \institution{Key Laboratory of Intelligent Computing and Information Processing, Ministry of
Education, School of Computer Science \& School of Cyberspace Security, Xiangtan University}
  \streetaddress{30 Shuangqing Rd}
  \city{Xiangtan}
  \state{Hunan}
  \country{China}}
\email{shwang@smail.xtu.edu.cn}

\author{Huanrong Tang}
\authornote{Corresponding author.}
\affiliation{%
  \institution{Key Laboratory of Intelligent Computing and Information Processing, Ministry of
Education, School of Computer Science \& School of Cyberspace Security, Xiangtan University}
  \streetaddress{30 Shuangqing Rd}
  \city{Xiangtan}
  \state{Hunan}
  \country{China}}
\email{tanghuanrong@126.com}
  
\author{Jianquan Ouyang}
\affiliation{%
  \institution{Key Laboratory of Intelligent Computing and Information Processing, Ministry of
Education, School of Computer Science \& School of Cyberspace Security, Xiangtan University}
  \streetaddress{30 Shuangqing Rd}
  \city{Xiangtan}
  \state{Hunan}
  \country{China}}
\email{oyjq@xtu.edu.cn}

\renewcommand{\shortauthors}{Trovato and Tobin, et al.}

\begin{abstract}
  This paper presents a neural architecture search method based on Transformer architecture, searching cross multihead attention computation ways for different number of encoder and decoder combinations. In order to search for neural network structures with better translation results, we considered perplexity as an auxiliary evaluation metric for the algorithm in addition to BLEU scores and iteratively improved each individual neural network within the population by a multi-objective genetic algorithm. Experimental results show that the neural network structures searched by the algorithm outperform all the baseline models, and that the introduction of the auxiliary evaluation metric can find better models than considering only the BLEU score as an evaluation metric.
\end{abstract}

\ccsdesc[500]{Computing methodologies~Genetic algorithms}
\ccsdesc[300]{Computing methodologies~Machine translation}

\keywords{genetic algorithm, Transformer, multi-objective}

\maketitle

\section{Introduction}
The Transformer \cite{vaswani2017attention} model has been used with great success in the application of neural machine translation. To further enhance its capabilities, this paper introduces the genetic algorithm-based neural architecture search \cite{liu2021survey} (GA-NAS) technique to the Transformer model, which breaks the fixed number and composition of encoders and decoders. To evaluate the translation effectiveness of the neural network, this paper uses two key metrics - the BLEU score \cite{papineni2002bleu} and perplexity. BLEU score, which measures the similarity between the predicted output and the reference translation, is used as the primary evaluation index. Perplexity, which measures the model's ability to predict the next word in a sequence, is used as an auxiliary evaluation index. To solve this problem, a multi-objective genetic algorithm is applied and denoted as MO-Trans in this paper.

\section{Proposed method}
\subsection{Framework of MO-Trans}
This paper uses MOEA/D \cite{zhang2007moea} as the algorithmic framework because it retains non-dominated individuals on the EP set in successive populations. Algorithm \ref{alg:1} shows the framework of the proposed MO-Trans method. The function $g_{}^{te}$ described in algorithm \ref{alg:1} is the Tchebyshev function defined in reference \citep{zhang2007moea}. Step 2 calculates the distance between any two weight vectors, and get $T$ weight vectors closest to each weight vector. The most time-consuming steps of algorithm 1 are step 3 and step 7 since the neural network corresponding to coding needs to be trained for evaluating each individual in population.

\begin{algorithm}[!h]
	\caption{Framework of MO-Trans}
	\label{alg:1}
	\KwIn{stop rule of algorithm; M neural network evaluation metrics; N uniformly distributed weight vectors $\boldsymbol{\lambda_1},\boldsymbol{\lambda_2},...,\boldsymbol{\lambda_N}$; The number of neighbors of each weight vector T.}
	\KwOut{set EP.}
	$EP\leftarrow \emptyset$\;
	for each $i=\{1,2,...,N\}$ let $B_{i}=\{i_1,i_2,...,i_T\}$, where $\boldsymbol{\lambda_{}^{i_1}},\boldsymbol{\lambda_{}^{i_2}},...,\boldsymbol{\lambda_{}^{i_T}}$ are the nearest T vectors to $\boldsymbol{\lambda_{}^{i}}$\;
	Initialize N individual transformer architectures according to the genetic coding strategy and train them to obtain m evaluation indicators,let $\boldsymbol{FV_i}=\boldsymbol{F}(\boldsymbol{x_i})$\;
	Initialize $z=\{z_1,z_2,...,z_m\}$\;
	\For{$i=1$ to $N$} {
	    Randomly select two indexes k and l from $B_i$, apply crossover and mutation operators to generate new individual y from $x_k,x_l$\;
	    Train individual y to obtain m evaluation metrics, for each $j=\{1,2,...,m\}$, if $z_j<f_j(y)$, let $z_j=f_j(y)$\;
	    for each $j\in B_i$, if $g_{}^{te}(y|\boldsymbol{\lambda_{}^{j}},z)\le g_{}^{te}(x_j|\boldsymbol{\lambda_{}^{j}},z)$, let $x_j= y$ and $\boldsymbol{FV_j}=\boldsymbol{F}(\boldsymbol{y})$\;
	    Remove all vectors in EP that are dominated by $\boldsymbol{F}(\boldsymbol{y})$, and add $\boldsymbol{F}(\boldsymbol{y})$ to EP if none of the vectors in EP dominate $\boldsymbol{F}(\boldsymbol{y})$\;
	}
    if the termination condition is not satisfied, back to line 5, else return EP.
\end{algorithm}

\subsection{Gene Encoding Strategy}
\noindent\textbf{Number of encoder/decoder blocks}\quad The baseline transformer model consists of six encoders and the same number of decoders, which are represented as an integer in our coding strategy.

\begin{figure}[ht]
\centering
\includegraphics[scale=0.4]{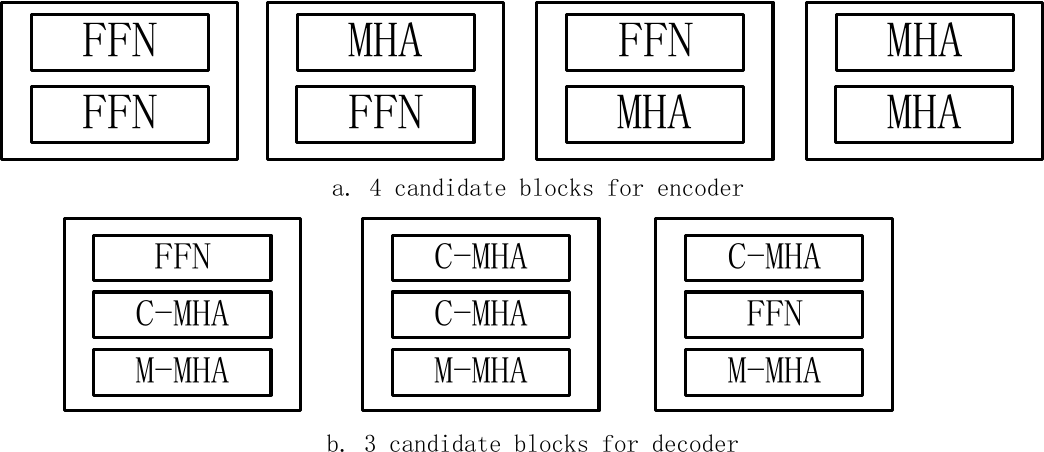}\caption{Searching space of combinations of blocks}
\label{fig:3}
\end{figure}

\noindent\textbf{Details of blocks}\quad  We have borrowed from Trans-GA \citep{feng2021evolving} the way the encoding and decoding blocks are composed, as shown in Figure \ref{fig:3}a and Figure \ref{fig:3}b, there exists 4 candidate blocks for encoder and 3 candidate blocks for decoder, that the M-MHA rectangle denotes the masked multihead attention layer and the C-MHA rectangle denotes the cross multihead attention layer. It can be noted that without a C-MHA layer, the decoder block will not compute the information from the encoder block, so unlike Trans-GA, this paper does not search for decoder blocks without a C-MHA layer. In addition to encoding the layer types, each MHA layer requires an integer to represent the number of heads and each FFN layer requires an integer to represent the dimensions.

\begin{figure}[ht]
\centering
\includegraphics[scale=0.5]{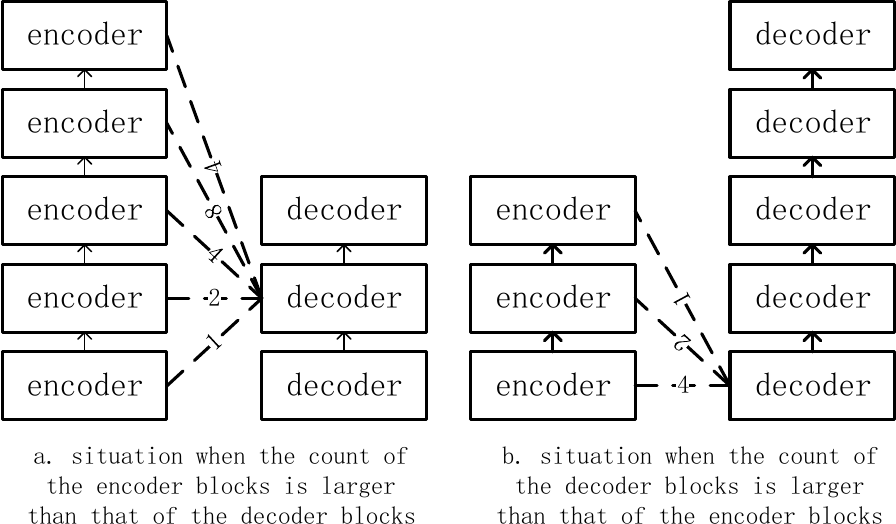}\caption{Cases in different numbers of encoder blocks and decoder blocks}
\label{fig:4}
\end{figure}

\noindent\textbf{Cross way}\quad The matrix $query$ of decoder blocks and the $key,value$ of encoder blocks are needed to compute multihead attention. In the baseline transformer model \cite{vaswani2017attention}, all decoder blocks are connected to the last encoder block to compute C-MHA. since the bottom encoder block tends to learn more syntax, and the top encoder block tends to learn more semantics, this paper prefers to connect a decoder block to the encoder block located close to it. In order not to waste the results of each encoder block, the last decoder block is connected directly to the last encoder block, but other decoder blocks just have a higher probability to connect to the encoder block which is near it, the nearer the higher. When the number of encoder blocks and decoder blocks is the same, this is a one-to-one relationship, when they are not the same, the situation may be somewhat different. As shown in figure \ref{fig:4}, whether there are more encoder blocks or more decoder blocks, the last encoder block must be connected to the last decoder block. As for the other blocks, when there are more encoder blocks, the dashed line shows that the two blocks with the same distance from the top have the highest probability weight of being connected to each other, with the weight being reduced by half for each additional distance, while when there are more decoder blocks, the situation is similar, with the two blocks with the same distance from the bottom being the most likely to be connected to each other, with all the extra decoder blocks being connected to the last encoder block. 

As mentioned above, a transformer architecture in searching space could be encoded as: 

$\{ne,[te,p1,p2]\times ne,nd,[td,p1,p2,p3,ce]
\times nd\}$,

which $ne$ and $nd$ denotes the number of encoder blocks and decoder blocks respectively, integer $te$ and $td$ denotes the type of candidate encoder block ranges $[1,4]$ and candidate decoder block ranges $[1,3]$ respectively, $p1,p2,p3$ denotes the number of heads in the MHA layer or the dimension in the FFN layer, integer $ce$ only uses in the decoder block which ranges $[1,ne]$ indicates which encoder block to compute cross multihead attention with. If set $ne=nd=6$ and both the number of heads in the MHA layer and the dimension in the FFN layer have 2 possible values, the size of searching space will reach $2.5\times 10^{19}$. 

\subsection{Genetic operators}
During population initialisation, all parameters except $ce$ are chosen from a range of uniform distributions. The genetic operator is described below.

\begin{figure}[ht]
\centering
\includegraphics[scale=0.4]{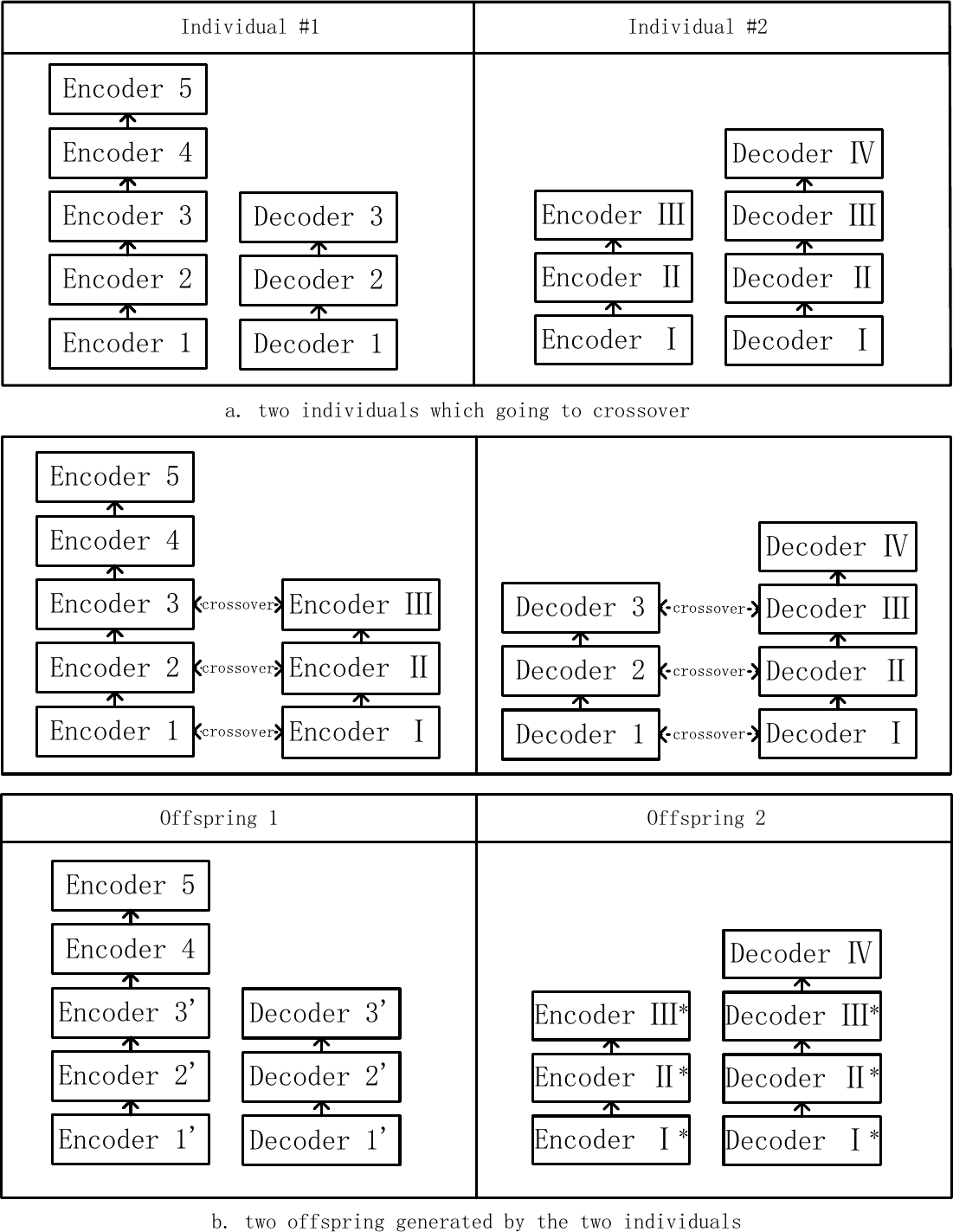}\caption{Illustration of crossover between individuals}
\label{fig:5}
\end{figure}

\noindent\textbf{Crossover}\quad This paper applies the idea of variable-length coding similar to EvoCNN \cite{sun2019evolving}. As shown in Figure \ref{fig:5}a and Figure \ref{fig:5}b, crossover operation between two individuals is encoder block to encoder block and decoder block to decoder block, the blocks numbered with Arabic numerals are from individual \#1 and those numbered with Roman numerals are from individual \#2. From Figure \ref{fig:5}b note that only the minimum number of the encoder blocks or the decoder blocks pairs will crossover, while the extra blocks will be put in the original position, so the parameter $ce$ does not need to be changed.

\noindent\textbf{Mutation}\quad these operations are available in mutation:

$\bullet$ Add an encoder/decoder block if the number will not exceed the upper bound.

$\bullet$ Drop an encoder/decoder block if the number will not below the lower bound.

$\bullet$ Alter the candidate type of an encoder/decoder block.

$\bullet$ Alter the number of heads in an MHA layer or the dimension in a FFN layer.

$\bullet$ Change the connection object encoder block from a decoder block.

\section{Experiment}
\subsection{Setup}
The experiments were conducted using the dataset Multi30k \citep{elliott2016multi30k}. The parameter $M$ used in Algorithm 1 was 2, and the two neural network metrics were perplexity and BLEU score, respectively. Since perplexity is generally considered to be negatively correlated with translation performance, the optimization objective was set to $(100-BLEU score, k\times perplexity)$, where $k$ is an adjustable parameter. The probabilities of crossover and mutation were set to 0.92 and 0.15 respectively. The number of heads in the MHA layer was chosen from the set $\{4,8\}$, the FFN dimension was chosen from the set $\{512,1024\}$, and the block sizes of the encoder and decoder ranged from $[3,7]$. The number of generations and the number of individuals in each generation were set to 15, with each individual running 10 epochs. Other parameters were set and initialised in essentially the same way as in \cite{vaswani2017attention}. The baseline model is the base Transformer model with an FFN dimension of 512. Both the individual population and baseline models introduce an early stop mechanism during training, where training is stopped if a lower loss value is not reached on the validation set within 2 epochs. The embedding size of all individual neural networks and baseline models was set to 512. The environment used for each experiment was an Nvidia Geforce RTX 3090 card.

\subsection{Results}
Tables \ref{tab:ende} and \ref{tab:deen} show the results of the comparison between English-German and German-English translations on the dataset for the baseline model and MO-Trans, respectively. In the table, \#E is the number of encoders, \#D is the number of decoders and \#Para is the number of parameters of the model. It can be observed that the algorithm searches for network structures with significantly better BLEU scores than the baseline model of all sizes. Noting that the algorithm will only consider the BLEU score as a single evaluation metric when k=0, and that the network architectures with the best translation results are obtained at k=0.75 and k=0.5 respectively, the introduction of perplexity as a secondary evaluation metric can in fact find a better network architecture than using a single evaluation metric.

\begin{figure}[ht]
\centering
\includegraphics[scale=0.35]{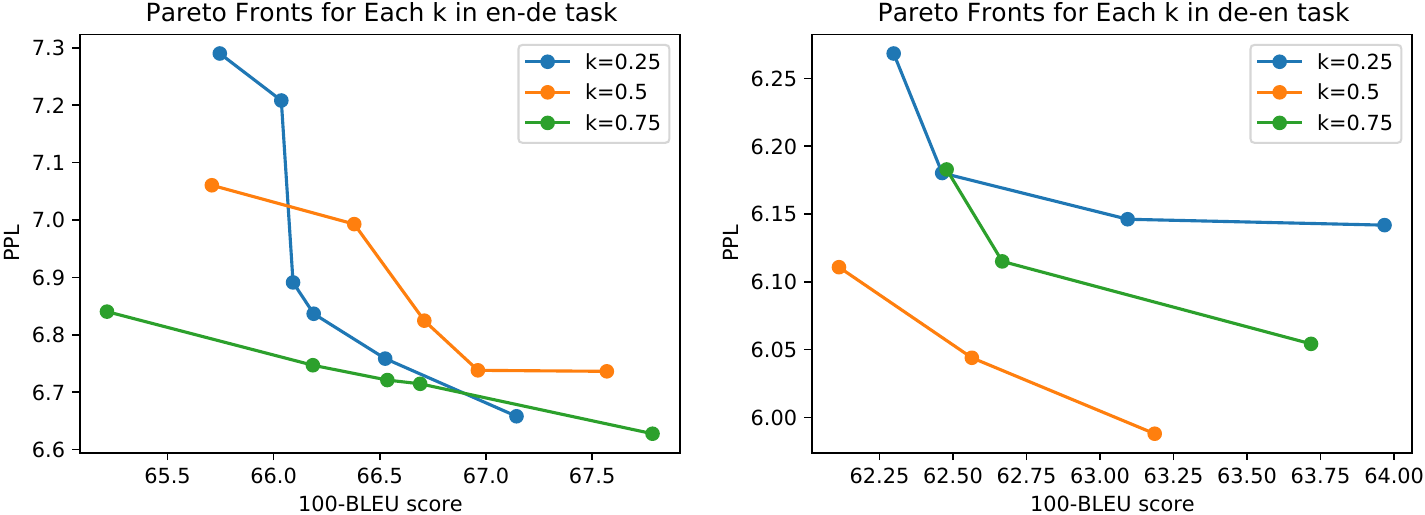}\caption{Illustration of the Pareto front of the population when the MO-Trans algorithm finished running}
\label{fig:6}
\end{figure}

\begin{figure}[ht]
\centering
\includegraphics[scale=0.38]{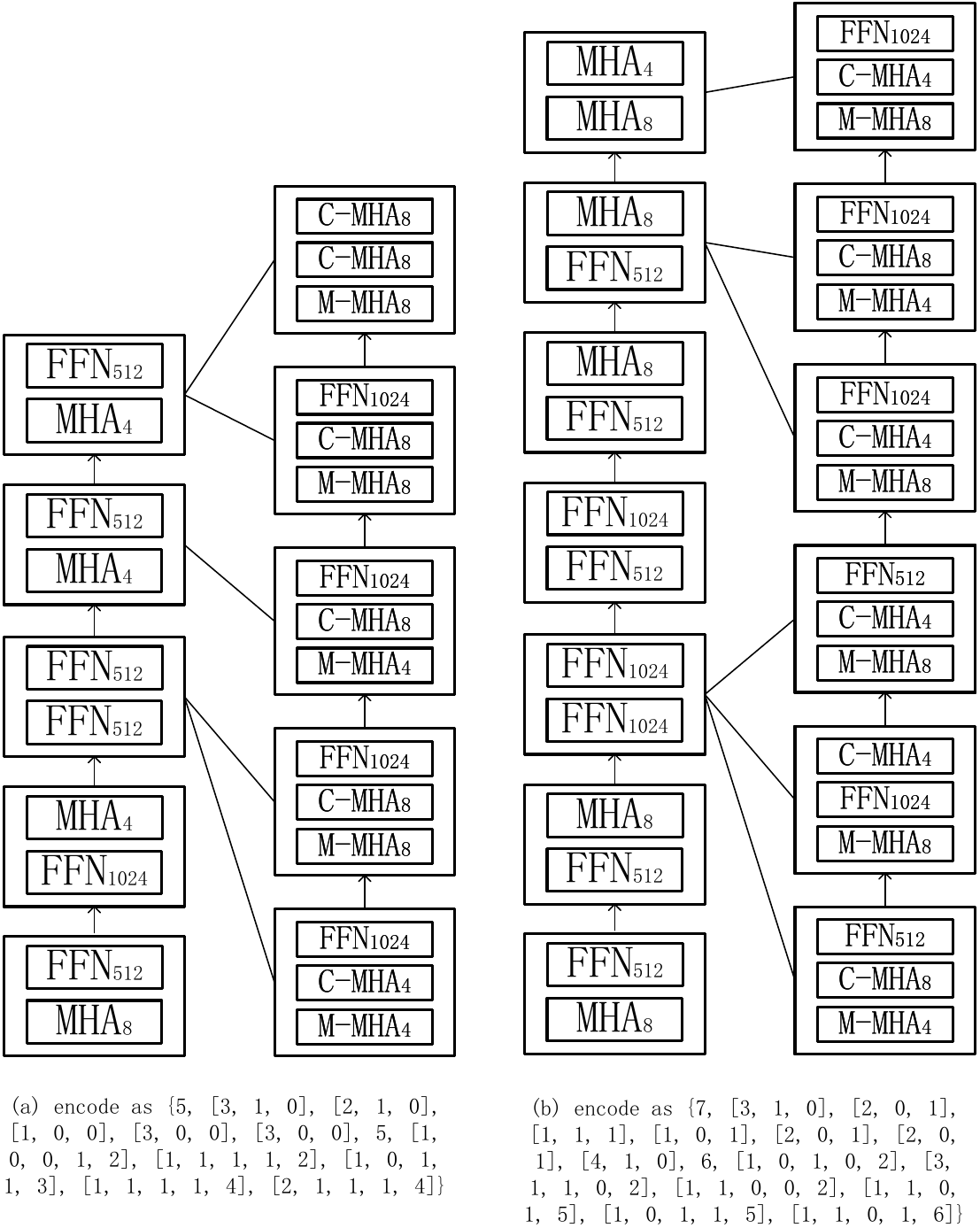}\caption{Schematic of the best model searched by the MO-Trans algorithm}
\label{fig:7}
\end{figure}

The Pareto front of the population is shown in Figure \ref{fig:6}. Figure \ref{fig:7}(a)(b) shows the best neural network structures found in the en-de and de-en tasks, respectively (their BLEU scores are highlighted in bold in Tables \ref{tab:ende} and \ref{tab:deen}). For the FFN layer, the value in the lower right corner represents the dimension, while for the MHA layer, the value in the lower right corner represents the number of attention heads.

\begin{table}
  \caption{Performance of the baseline model and MO-Trans on the en-de translation task for the Multi30k dataset}
  \label{tab:ende}
  \begin{tabular}{cccccc}
    \toprule
    \textbf{model} &\textbf{\# of E}&\textbf{\# of D} &\textbf{\# of Para} &\textbf{BLEU} \\
    \midrule
    baseline & 3 & 3 & 27.1M & 32.91\\
    baseline & 4 & 4 & 31.3M & 33.25\\
    baseline & 5 & 5 & 35.5M & 32.16\\
    baseline & 6 & 6 & 39.7M & 32.62\\
    baseline & 7 & 7 & 44.0M & 31.77\\
    $\rm MO\_Trans_{\ k=0}$ & 5 & 5 & 39.2M & 34.21\\ 
    $\rm MO\_Trans_{\ k=0.25}$ & 6 & 6 & 42.9M & 34.25\\ 
    $\rm MO\_Trans_{\ k=0.5}$ & 5 & 6 & 41.8M & 34.29\\ 
    $\rm MO\_Trans_{\ k=0.75}$ & 5 & 5 & 38.2M & \textbf{34.79}\\ 
  \bottomrule
\end{tabular}
\end{table}

\begin{table}
  \caption{Performance of the baseline model and MO-Trans on the de-en translation task for the Multi30k dataset}
  \label{tab:deen}
  \begin{tabular}{ccccc}
    \toprule
    \textbf{model} &\textbf{\# of E}&\textbf{\# of D} &\textbf{\# of Para} &\textbf{BLEU}\\
    \midrule
    baseline & 3 & 3 & 26.2M & 36.36\\
    baseline & 4 & 4 & 30.4M & 36.27\\
    baseline & 5 & 5 & 34.6M & 36.81\\
    baseline & 6 & 6 & 38.8M & 35.20\\
    baseline & 7 & 7 & 43.0M & 36.73\\
    $\rm MO\_Trans_{\ k=0}$ & 7 & 4 & 36.7M & 37.74\\ 
    $\rm MO\_Trans_{\ k=0.25}$ & 4 & 6 & 36.7M & 37.70\\ 
    $\rm MO\_Trans_{\ k=0.5}$ & 7 & 6 & 43.6M & \textbf{37.89}\\ 
    $\rm MO\_Trans_{\ k=0.75}$ & 5 & 6 & 39.9M & 37.52\\ 
  \bottomrule
\end{tabular}
\end{table}

\section{Conclusion}
This paper presents a neural architecture search method based on Transformer model considering multiple evaluation metrics for machine translation tasks, MO-Trans. Experimental results demonstrate that the introduction of a search for cross multihead attention computation methods and the consideration of auxiliary evaluation metrics boost the effectiveness of translation. The ideas in this paper would be helpful in designing a better Transformer model. \footnote{ The reference code of this paper is published on https://github.com/ra225/MO-Trans.}

\begin{acks}
This research has been supported by Key Projects
of the Ministry of Science and Technology of the People Republic of China
(No.2020YFC0832405).
\end{acks}
\bibliographystyle{ACM-Reference-Format}
\bibliography{sample-base}


\end{document}